# The Latent Structure of Dictionaries

Philippe Vincent-Lamarre[1,2], Alexandre Blondin Massé[1], Marcos Lopes[3], Mélanie Lord[1], Odile Marcotte[1], Stevan Harnad[1,4]

**1** Université du Québec à Montréal, **2** Université d'Ottawa, **3** University of São Paulo (USP), **4** University of Southampton

**ABSTRACT:** How many words – and which ones – are sufficient to define all other words? When dictionaries are analyzed as directed graphs with links from defining words to defined words, they reveal a latent structure. Recursively removing all words that are reachable by definition but that do not define any further words reduces the dictionary to a *Kernel* of about 10%. This is still not the smallest number of words that can define all the rest. About 75% of the Kernel turns out to be its *Core*, a "Strongly Connected Subset" of words with a definitional path to and from any pair of its words and no word's definition depending on a word outside the set. But the Core cannot define all the rest of the dictionary. The 25% of the Kernel surrounding the Core consists of small strongly connected subsets of words: the *Satellites*. The size of the smallest set of words that can define all the rest – the graph's "minimum feedback vertex set" or *MinSet* – is about 1% of the dictionary,15% of the Kernel, and half-Core/half-Satellite. But every dictionary has a huge number of MinSets. The Core words are learned earlier, more frequent, and less concrete than the Satellites, which in turn are learned earlier and more frequent but more concrete than the rest of the Dictionary. In principle, only one MinSet's words would need to be *grounded* through the sensorimotor capacity to recognize and categorize their referents. In a dual-code sensorimotor/symbolic model of the mental lexicon, the symbolic code could do all the rest via re-combinatory definition.

**The Representation of Meaning.** One can argue that the set of all the written words of a language constitutes the biggest and richest digital database on the planet. Numbers and algorithms are just special cases of words and sentences, so they are all part of that same global verbal database. Analog images are not words, but even their digitized versions only become tractable once they are sufficiently tagged with verbal descriptions. So in the end it all comes down to words. But how are the meanings of words represented? There are two prominent representations of word meaning: one is in our external dictionaries and the other is in our brains: our "mental lexicon." How are the two related?

**The Symbol Grounding Problem.** We consult a dictionary in order to learn the meaning of a word whose meaning we do not yet already know. Its meaning is not yet in our mental lexicon. The dictionary conveys that meaning to us through a definition consisting of further words, whose meanings we already know. If a definition contains words whose meanings we do not yet know, we can look up their definitions too. But it is clear that meaning cannot be dictionary look-up all the way down. *The meanings of some words, at least, have to be learned by some means other than dictionary look-up, otherwise word meaning is ungrounded*: just strings of meaningless symbols (defining words) pointing to meaningless symbols (defined words). This is the "symbol grounding problem" (Harnad 1990).

This paper addresses the question of *how many words* – and *which words* – have to be learned (grounded) by means other than dictionary look-up so that all the rest of the words in the dictionary can be defined either directly, using only combinations of those grounded words, or, recursively, using further words that can themselves be defined using solely those grounded words. Let us call those grounded words in our mental lexicon – the ones sufficient to define all the others – a "Grounding Set."

**Category Learning.** The process of word grounding itself is the subject of a growing body of ongoing work on the sensorimotor learning of categories, by people as well as by computational models (Harnad 2005; De Vega, Glenberg & Graesser 2008; Ashby & Maddox 2011; Meteyard, Cuadrado, Bahrami & Vigliocco 2012; Pezzulo, Barsalou, Cangelosi, Fischer, McRae & Spivey 2012; Blondin Massé, Harnad, Picard & St-Louis 2013; Kang 2014; Maier, Glage, Hohlfeld & Rahman 2014). Here we just note that almost all the words in any dictionary (nouns, verbs, adjectives and adverbs) are "content" words,[1] meaning that they are the names of *categories* (objects, individuals, kinds, states, actions, events, properties, relations) of various degrees of abstractness. The more concrete of these categories, and hence the words that name them, can be learned directly through trial-and-error sensorimotor experience, guided by feedback that indicates whether an attempted categorization was correct or incorrect. The successful result of this learning is a sensorimotor category representation – that is, a feature-detector that enables the learner to categorize sensory inputs correctly, identifying them with the right category name (Ashby & Maddox 2011; Folstein, Palmeri, Van Gulick & Gauthier 2015; Hammer, Sloutsky & Grill-Spector 2015). A grounding set composed of such experientially grounded words would then be enough (in principle, though not necessarily in practice) to allow the meaning of all further words to be learned through verbal definition alone. It is only the symbolic module of such a dual-code sensorimotor/symbolic system for representing word meaning (Paivio 2014) that is the object of study in this paper. But the underlying assumption is that the symbolic code is grounded in the sensorimotor code.

**Expressive Power.** Perhaps the most remarkable and powerful feature of natural language is the fact that *it can say anything and everything that can be said* (Katz 1978). There exists no language in which you can say this, but not that. (Pick a pair of languages and try it out.) Word-for-word translation may not work: you may not be able to say everything in the same number of words, equally succinctly, or equally elegantly, in the same *form*. But you will always be able to translate in paraphrase the *propositional content* of anything and everything that can be said in any one language into any other language. (If you think that may still leave out anything that can be said, just say it in any language at all and it will prove to be sayable in all the others too; Steklis & Harnad 1976.)

One counter-intuition about this is that the language may lack the words: its vocabulary may be insufficient: How can you explain quantum mechanics in the language of isolated Amazonian hunter-gatherers? But one can ask the very same question about how you can explain it to an American 6-year-old – or, for that matter, to an eighteenth century physicist. And the banal answer is that it takes time, and a lot of words, to explain – but you can always do it, in any language. Where do all those missing words come from, if not from the same language? We coin (i.e., lexicalize) words all the time, as they are needed, but we are coining them *within* the same language; it does not become a different language every time we add a new word. Nor are most of the new words we coin labels for unique new experiences, like names for new colors (e.g., "ochre") or new odors ("acetic") that you have to see or smell directly at first hand in order to know what their names refer to.

---

[1] Content or Open Class words are growing in all spoken languages all the time. In contrast, Function or Closed Class words like if, off, is, or his are few and fixed, with mostly a formal or syntactic function: Our study considers only content words. Definitions are treated as unordered strings of content words, ignoring function words, syntax and polysemy (i.e, multiple meanings, of which we use only the first and most common meaning for each word-form).

Consider the German word "*Schadenfreude*" for example. There happens to be no single word for this in English. It means "feeling glee at another's misfortune." English is highly assimilative, so instead of bothering to coin a new English word (say, "misfortune-glee," or, more latinately, "malfelicity") whose definition is "glee at another's misfortune," English has simply adopted *Schadenfreude* as part of its own lexicon. All it needed was to be defined, and then it could be added to the English dictionary. The shapes of words themselves are arbitrary, after all, as Saussure (1911/1972) noted: words do not resemble the things they refer to.

So what it is that gives English or any language its limitless expressive power is *its capacity to define anything with words*. But is this defining power really limitless? First, we have already skipped over one special case that eludes language's grasp, and that is new sensations that you have to experience at first hand in order to know what they are – hence to understand what any word referring to them means. But even if we set aside words for new sensations, what about other words, like *Schadenfreude*? That does not refer to a new sensory experience. We understand what it refers to because we understand what the words "glee at another's misfortune" refer to. That definition is itself a *combination* of words; we have to understand those words in order to understand the definition. If we don't understand some of the words, we can of course look up their definitions too – but as we have noted, it cannot be dictionary look-ups all the way down! The meanings of some words, at least (e.g., "glee") need to have been grounded in direct experience, whereas others (e.g., "another" or "misfortune") may be grounded in the meaning of words that are grounded in the meaning of words… that are grounded in direct experience.

**Direct Sensorimotor Grounding.** How the meaning of a word referring to a sensation like "glee" can be grounded in direct experience is fairly straightforward: It's much the same as teaching the meaning of "ochre" or "acetic": "Look (sniff): that's ochre (acetic) and look (sniff) that's not." "Glee" is likewise a category of perceptual experience. To teach someone which experience "glee" is, you need to point to examples that are members of the category "glee" and examples that are not: "Look, that's glee" – pointing to someone who looks and acts and has reason to feel gleeful - and "Look, that's not glee" – pointing[2] to someone who looks and acts and has reason to feel ungleeful (Harnad 2005).

---

[2] Wittgenstein had some cautions about the possibility of grounding words for private experiences because there would be no basis for correcting errors. But thanks to our "mirror neurons" and our "mind-reading" capacity we are adept at inferring most private experiences from their accompanying public behavior, and reasonable agreement on word meaning can be reached on the basis of common experience together with these observable behavioral correlates (Apperley 2010). Because of the "other-minds problem" – i.e., because the only

What about the categories denoted by the words “another” and “misfortune”? These are not direct, concrete sensory categories, but they still have examples in our direct sensorimotor experience: “That’s you” and “that’s another” (i.e., someone else). “That’s good fortune” and “that’s misfortune.” But it is more likely that higher-order, more abstract categories like these would be grounded in verbal definitions composed of words that each name already grounded categories, rather than being grounded in direct sensorimotor experience (Summers 1988; Aitchison 2012; Huang & Eslami 2013; Nesi 2014).

**Dictionary Grounding**. This brings us to the question that is being addressed in this paper: A dictionary provides an (approximate) definition for every word in the language. Apart from a small, fixed set of words whose role is mainly syntactic (“function words,” e.g. articles, particles, conjunctions), all the rest of the words in the dictionary are the names of categories (“content words,” i.e. nouns, verbs, adjectives, adverbs). *How many content words (i) – and which ones (ii) – need to be grounded already so that all the rest can be learned from definitions composed only out of those grounded words?* We think the answer casts some light on how the meaning of words is represented – externally, in dictionaries, and internally, in our mental lexicon – as well as on the evolutionary origin and adaptive function of language for our species (Cangelosi & Parisi 2012; Blondin Massé et al. 2013).

**Synopsis of Findings.** Before we describe in detail what we did, and how, here is a synopsis of what we found: When dictionaries are represented as graphs, with arrows from each defining word to each defined word, their graphs reveal a latent structure that had not previously been identified or reported, as far as we know. Dictionaries have a special subset of words (about 10%) that we have called their “Kernel” words. **Figure 1** illustrates a dictionary and its latent structure using a tiny mini-dictionary graph (Picard, Lord, Blondin Massé, Marcotte, Lopes & Harnad 2013) derived from an online definition game that will be described later. A full-sized dictionary has the same latent structure as this mini-dictionary, but with a much higher proportion of the words (90%) lying outside the Kernel. **Table 1** and **Figure 2** show the proportions for a full-size dictionary.

The dictionary’s Kernel is unique, and its words can define the remaining 90% of the dictionary (the “Rest”). The Kernel is hence a grounding set. But it is not the *smallest* grounding set. This smallest grounding subset of the Kernel – which we have called the “Minimal Grounding Set” (“MinSet”) – turns out to be much smaller than the Kernel (about 15% of the Kernel and 1% of the whole dictionary), but it is not unique: The Kernel contains a huge number of different MinSets. Each of these is of the same minimum size and each is able to define all the other words in the dictionary.

The Kernel also turns out to have further latent structure: About 75% of the Kernel consists of a very big “strongly connected component” (SCC: a subset of words within which there is a definitional path to and from every pair of words but no incoming definitional links from

---

experiences you can have are your own – there is no way to know for sure whether private experiences accompanied by the same public behavior are indeed identical experiences. These subtleties do not enter into the analyses we are doing in this paper. Word meaning is in any case not exact but approximate in all fields other than formal mathematics and logic. Even observable, empirical categories can only be defined or described provisionally and approximately: Like a picture or an object, an experience is always worth more than a thousand (or any number) of words (Harnad 1987).

outside the SCC). We call this the Kernel's (and hence the entire dictionary's) "Core." The remaining 25% of the Kernel surrounding the Core consists of many tiny strongly connected subsets, which we call the Core's "Satellites." It turns out that each MinSet is part-Core and part Satellite.

The words in these distinct latent structures also turn out to differ in their psycholinguistic properties: As we go deeper into the dictionary, from the 90% Rest to the 10% Kernel, to the Satellites (1-4%) surrounding the Kernel's Core, to the Core itself (6-9%), the words turn out on average to be more frequently used (orally and in writing) and to have been learned at a younger age. This is reflected in a gradient within the Satellite layer: the shorter a Satellite word's definitional distance (the number of definitional steps to reach it) from the Core, the more frequently it is used and the earlier it was learned. The average concreteness of the words within the Core and the 90% of words that are outside the Kernel (i.e., the Rest) is about the same. Within the Satellite layer in between them, however, Kernel words become more concrete the greater their definitional distance outward from the Core. There is also a (much weaker) definitional distance gradient from the Kernel outward into the 90% Rest of the dictionary for age and concreteness, but not for frequency. We will now describe how this latent structure was discovered.

**Control Vocabularies.** Our investigation began with two small, special dictionaries – the *Cambridge International Dictionary of English* (47,147 words; Procter 1995; henceforth *Cambridge*) and the *Longman Dictionary of Contemporary English* (69,223 words; Procter 1978; henceforth *Longman*) (Table 1). These two dictionaries were created especially for people with limited English vocabularies, such as non-native speakers; all words are defined using only a "control" vocabulary of 2000 words that users are likely to know already. Our objective was to analyze each dictionary as a directed graph (*digraph*) in which there is a directional link from each defining word to each defined word. Each word in the dictionary should be reachable, either directly or indirectly, via definitions composed of the 2000-word control vocabulary.

A direct analysis of the graphs of *Longman* and *Cambridge*, however, revealed that their underlying control-vocabulary principle was not faithfully followed: There turned out to be words in each dictionary that were not defined using only the 2000-word "control" vocabulary, and there were also words that were not defined at all. So we decided to use each dictionary's digraph (a directed graph with arrows pointing from the words in each definition to the word they define) to work backward in order to see if we could generate a genuine control vocabulary out of which all the other words could be defined. (We first removed all undefined words.)

**Dictionaries as Graphs.** Dictionaries can be represented as directed graphs $D = (V, A)$ (*digraphs)*. The *vertices* are words and the *arcs* connect defining words to defined words, i.e. there is an arc from word *u* to word *v* if u is a word in the definition of v. Moreover, in a complete dictionary, every word is defined by at least one word, so we assume that there is no word without an incoming arc. A *path* is a sequence $(v_1, v_2, \ldots, v_k)$ of vertices such that $(v_i, v_{i+1})$ is an arc for $i = 1,2, \ldots, k-1$. A *circuit* is a path starting and ending at the same vertex. A graph is called *acyclic* if it does not contain any circuits.

**Grounding Sets.** Let $U \subseteq V$ be any subset of words and let *u* be some given word. We are interested in computing all words that can be learned through definitions composed only of words in $U$. This can be stated recursively as follows: We say that u is *definable* from $U$ if all predecessors of $u$ either belong to $U$ or are definable from $U$. The set of words that can be defined from $U$ is denoted by $Def\ (U)$. In particular, if $Def(U) \cup U = V$, then U is called a *grounding set* of $D$. Intuitively, a set $U$ is a grounding set if, provided we already know the meaning of each word in $U$, we can learn the meaning of all the remaining words just by looking up the definitions of the unknown words (in the right order).

Grounding sets are equivalent to well-known sets in graph theory called *feedback vertex sets* (Festa, Pardalos & Resende (1999). These are sets of vertices $U$ that *cover* all circuits, i.e. for any circuit *c*, there is at least one word of *c* belonging to $U$. It is rather easy to see this. On the one hand, if there exists a circuit of unknown words, then there is no way to learn any of them by definition alone. On the other hand, if every circuit is covered, then the graph of unknown words is acyclic, which means that the meaning of at least one word can be learned – a word having no unknown predecessor (Blondin Massé et al 2008)).

Clearly, every dictionary D has many grounding sets. For example, the set of all words in D is itself a grounding set. But how small can grounding sets be? In other words, what is the smallest number of words you need to know already in order to be able to learn the meaning of all the remaining words in D through definition alone? These are the Minimal Grounding Sets (*MinSets)* mentioned earlier (Fomin, Gaspers, Pyatkin & Razgon 2008). It is already known that finding a minimum feedback vertex set in a general digraph is NP-hard (Karp, 1972), which implies that finding MinSets is also NP-hard. Hence, it is highly unlikely that one will ever find an algorithm that solves the general problem without taking an exponentially long time. However, because some real dictionary graphs are relatively small and also seem to be structured in a favorable way, our algorithms are able to compute their MinSets.

**Kernel.** As a first step, we observed that in all dictionaries analyzed so far there exist many words that are never used in any definition. These words can be removed without changing the MinSets. This reduction can be done iteratively until no further word can be removed without leaving any word undefinable from the rest. The resulting subgraph is what we called the dictionary's *(grounding) Kernel*. Each dictionary's Kernel is unique, in the sense that every dictionary has one and only one Kernel. The Kernels of our two small dictionaries, *Longman* and *Cambridge,* turned out to amount to 8% and 7% of the dictionary as a whole, respectively. We have since extended the analysis to two larger dictionaries, *Merriam-Webster* (248,466 words; Webster 2006; henceforth *Webster*) and *WordNet* (132,477 words; Fellbaum 2010) whose Kernels are both 12% of the dictionary as a whole (**Table 1**).

**Core and Satellites.** Next, since we are dealing with directed graphs, we can subdivide the words according to their *strongly connected components*. Two words *u* and *v* are *strongly connected* if there exists a path from *u* to *v* as well as a path from *v* to *u*. *Strongly Connected Components* (SCCs) are hence maximal sets of words with a definitional path to and from any pair of their words. There is a well-known algorithm in graph theory that computes all the SCCs very efficiently (Tarjan 1972). *Sources* are SCCs in which no word's definition depends on a word outside the SCC (no incoming arcs). The Kernel of each of the four dictionary graphs turns out to contain an SCC much larger than all the others. One would intuitively expect the Kernel's Core (as the union of Sources) to be that largest SCC.

And so it is in two of the four dictionaries we analyzed. But because of the algorithm we used in preprocessing our dictionaries, in the other two dictionaries the Core consists of the largest SCC plus a few extra (small) SCCs. We think those small extra SCCs are just an artifact of the preprocessing. In any case, for each of the four dictionaries, the Core amounts to 65%-90% of the Kernel or about 6.5%-9.0% of the dictionary as a whole. The SCCs that are inside the Kernel but outside the Core are called *Satellites*; collectively they make up the remaining 10%-35% of the Kernel or about 1.0%-3.5% of the whole dictionary. *Satellites*.

**Definitional Distance from the Kernel: the *K*-Hierarchy.** Another potentially informative graph-theoretic property is the "definitional distance" of any given word from the Kernel or from the Core in terms of the number of arcs separating them. We define these two distance hierarchies as follows. First, for the Kernel hierarchy, suppose *K* is the Kernel of a dictionary graph *D*. Then, for any word *u*, we define its distance recursively as follows:

1. $dist(u) = 0$, if *u* is in *K*;
2. $dist(u) = 1 + max\{dist(v) : v \text{ is a predecessor of } u\}$, otherwise.

In other words, to compute the distance between K, as origin, and any word *u* in the rest of D, we compute the distances of all words defining *u* and add one. This distance is well defined, because *K* is a grounding set of *D* and hence the procedure cannot cycle because every circuit is covered. The mapping that relates every word to its distance from the Kernel is called the *K-hierarchy*.

**Definitional Distance from the Core: the C-Hierarchy.** The second metric is slightly more complicated but based on the same idea. Let *D* be the directed graph of a dictionary, and *D'* be the graph obtained from *D* by merging each strongly connected component (SCC) into a single vertex. The resulting graph is acyclic. We can then compute the distance of any word from the Core (the vertex corresponding to the biggest of the merged strongly connected components of the Kernel) as follows:

1. $dist(u) = 0$, if *u* is in a source vertex of *D'*;
2. $dist(u) = 1 + max\{dist(v) :$ $v \text{ is a predecessor of } w \text{ for some } w \text{ in the same SCC as } u\}$, otherwise.

The words in the merged vertices of the Core have no predecessor and constitute the origin of the C-hierarchy. Like the K-hierarchy, the C-hierarchy is well defined because *D'* is acyclic.

**MinSets.** We have computed the Kernel K, Core C, and Set of satellites $S$ as well as the K-hierarchy and the C-hierarchy for four English dictionaries: two smaller ones – (1) Longman's Dictionary of Contemporary English (Longman, 47,147 words), (2) Cambridge's International Dictionary of English (Cambridge, 69,223 words) – and two larger ones - (3) Merriam-Webster (Webster, 248,466 words), (4) WordNet (132,477 words). Because of polysemy (multiple meanings)[3], there can be more than one word with the same word-form

[3] Once the problem of polysemy is solved for both defined and defining words, the analysis described in this paper can be applied to each unique word/meaning pair instead of just to the first meaning of each defined word.

(lexeme). As an approximation, for each stemmatized word-form we used only the first (and most frequent) meaning for each part of speech of that word-form (noun, verb, adjective, adverb). (This reduced the total number of words by 53% for Cambridge, 49% for Longman, 37% for Webster and 65% for WordNet.) The sizes of their respective Kernels turned out to be between 8% of the whole dictionary for the smaller dictionaries and 12% for the larger dictionaries. The Kernel itself varied from 10% Satellite and 90% Core for the two small dictionaries to 35% Satellite and 65% Core for the two large dictionaries (**Table 1**).

As noted earlier, computing the MinSets is much more difficult than computing K, C, or S (in our current state of knowledge), because the problem is NP-hard. (Note that the most difficult part consists of computing MinSets for the Core, which can be further reduced by a few simple operations.) This problem can be modelled as an integer linear program whose constraints correspond to set-wise minimal circuits in the dictionary graph. The number of these constraints is huge but one can "add" constraints as they are needed. For the two "small" dictionaries (Longman and Cambridge), we were able to use CPLEX, a powerful optimizer, to compute a few MinSets (although not all of them, because there are a very large number of MinSets). For Webster and WordNet, the MinSets that we obtained after several days of computation were almost optimal.[4]

These analyses answered our first question about the *size* of the MinSet for these four dictionaries (373 and 452 words for the small dictionaries; 1396 and 1094 for the larger ones; about 1% for each dictionary). But because, unlike a dictionary's unique Kernel, its MinSets are not unique, a dictionary has a vast number of MinSets, all within the Kernel, all the same minimal size, but each one different in terms of which combination of Core and Satellite words it is composed of. The natural question to ask now is whether the words contained in these latent components of the dictionary, identified via their graph-theoretic properties – the MinSets, Core, Satellites, Kernel and the rest of the dictionary – differ from one another in any systematic way that might give a clue as to the function (if any) of the different latent structures identified by our analysis.

[4] This is yet another approximation in an analysis that necessitated many approximations: ignoring syntax and word order, using only the first meaning, and finding only something close to the MinSet for the biggest dictionaries. Despite all these approximations and potential sources of error, systematic and interpretable effects emerged from the data.

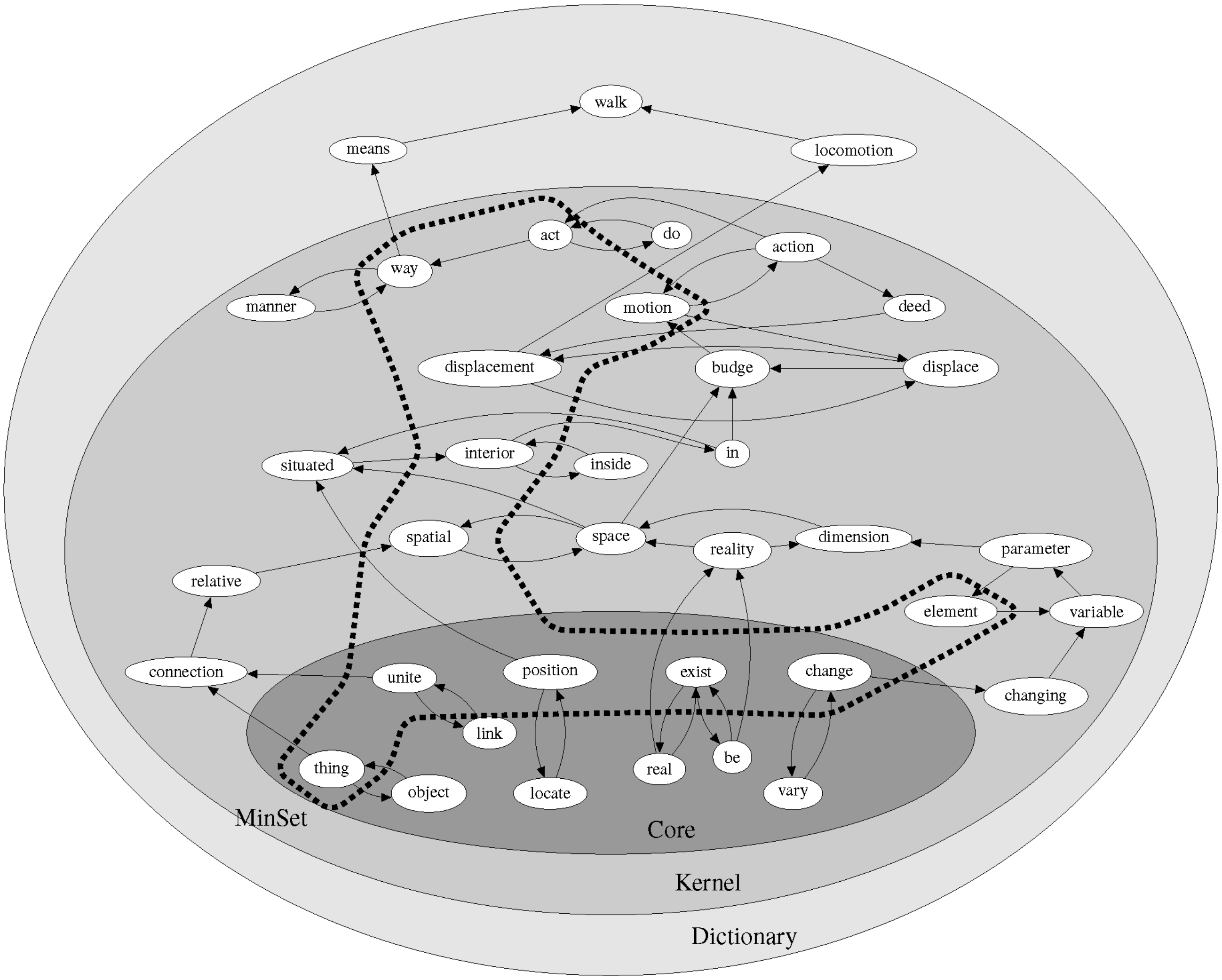

**Figure 1.** Illustration of a dictionary graph using data for a tiny (but complete) mini-dictionary (Picard et al 2013) generated by our dictionary game (see text for explanation). Arrows are from defining words to defined words. The entire mini-dictionary consists of just 32 words. Mini-dictionaries have all the latent structures of full-sized dictionaries. The smallest ellipse is the Core. The medium-sized ellipse is the Kernel. The part of the Kernel outside the Core is the Satellites. The part outside the Kernel is the Rest of the dictionary. There are many MinSets, all part-Core and part-Satellites (only one MinSet is shown here). In full-sized dictionaries the Rest is about 90% of the dictionary but in the game mini-dictionaries the Kernel is about 90% of the dictionary. The average Satellite-to-Core ratio in the Kernel for full-sized and mini-dictionaries is about the same (3/7) (see Table 1) but within MinSets this ratio is reversed (2/5 for full dictionaries and 5/2 for mini-dictionaries).

| | Cambridge | Longman | Webster | WordNet | Game dictionaries (average) |
|---|---|---|---|---|---|
| *Total word-meanings* | *47147* | *69223* | *248466* | *132477* | *182* |
| **First word-meanings** | **25132** | **31026** | **91388** | **85195** | 182 |
| **Rest** | **22891 (91%)** | **28700 (93%)** | **80433 (88%)** | **75393 (88%)** | 10.1 (7%) |
| **Kernel** | **2241 (9%)** | **2326 (8%)** | **10955 (12%)** | **9802 (12%)** | 171.7 (93%) |
| Satellites | 232 (1%) | 540 (2%) | 2978 (3%) | 3410 (4%) | 54.5 (29%) |
| Core | 2009 (8%) | 1786 (6%) | 7977 (9%) | 6392 (8%) | 117.2 (64%) |
| MinSets | 373 (1%) | 452 (1%) | 1396 (2%) | 1094 (1%) | 32.8 (18%) |
| Satellite-MinSets | 59 (16%) | 167 (37%) | 596 (43%) | 532 (49%) | 20.6 (63%) |
| Core-MinSets | 314 (84%) | 285 (63%) | 800 (57%) | 562 (51%) | 12.2 (37%) |

**Table 1.** Number and percentage of word-meanings for each latent structure in each of the four dictionaries used (plus averages for game-generated dictionaries). Based on using only the first word-meaning for each stemmatized part of speech wherever there are multiple meanings (hence multiple words).

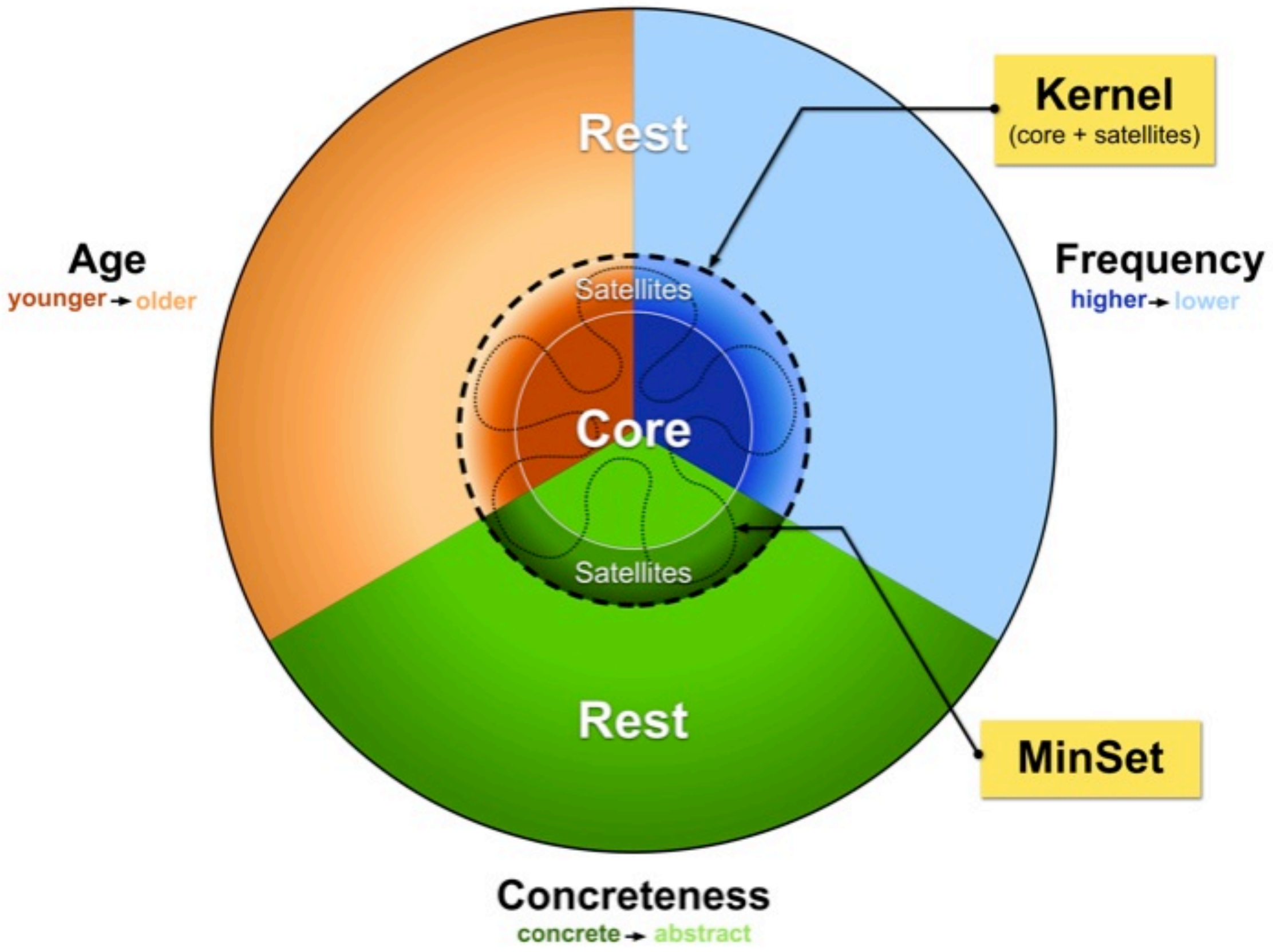

**Figure 2.** Overall pattern for average psycholinguistic differences (age of acquisition, concreteness, frequency) between words in latent structures revealed by the analysis of the dictionary digraph. Pattern is the same for all four dictionaries analyzed but image is not drawn to scale: for exact numbers and percentages see **Table 1** and **Figures 3** & **7**). (MinSets are part Core and part Satellite. Core + Satellites = Kernel [~10%]. Outside the Kernel is the Rest [~90%]). Core words are more frequent (blue) and learned younger (orange) than the Rest of the dictionary. Within the Kernel's Satellite layer, this difference increases gradually as definitional distance from the Core increases. Outside the Kernel, for age, the difference decreases gradually (but weakly) as definitional distance from the Kernel increases; frequency remains uniform. For concreteness (green), it is the Satellite layer that is more concrete than the Core. This difference increases gradually as definitional distance from the Core increases within the Satellite layer of the Kernel. Outside the Kernel, concreteness is at first equal to the Core and then increases gradually (but weakly) as definitional distance from the Kernel increases.

**Psycholinguistic Correlates of Dictionary Latent Structure.** A number of databases have been compiled that index various psycholinguistic properties of words (e.g., Wilson 1988). We used three of them: For word frequency, we used the $SUBTLEX_{US}$ Corpus, which has been found to be more reliable than the widely used Kučera and Francis (1967) word frequency norms (Brysbaert & New, 2009). Raw frequencies range from 1 to over 2 million, with an average of 669 and with about 1% of the values over 5000. For our goal of determining the average frequency for different sets of words, instead of using raw frequency, we used the Lg10WF metric (log10(FREQcount+1)) to reduce the effect of extreme values. For concreteness, the Brysbaert, Warriner & Kuperman (2014) concreteness ratings for 40,000 common English word lemmas were used. For age of acquisition, we used the Kuperman et al. (2012) age-of-acquisition ratings for 30,000 English words.

We tested whether the words in the latent components we identified in dictionary graphs differ systematically in frequency, concreteness or age of acquisition. Our overall pattern of findings (for all four dictionaries) is illustrated in **Figure 2**, which shows the latent structures of the dictionary: the 90% Rest and the 10% Kernel, and within it the Core surrounded by its Satellites. Shown also is one MinSet (just one of many); all MinSets are part Core and part Satellite.

Based on the data for word frequency (blue), concreteness (green) and age of acquisition (orange) from the psycholinguistic databases, the words in the Core for all four dictionaries are more frequent and learned younger than the Satellite words, which are in turn more frequent and younger than the Rest of the dictionary. The Satellites are more concrete than the Core or the Rest. The average values for each of the psycholinguistic variables in each of the latent substructures are shown in **Figure 3**. The pattern is the same for all four dictionaries. Because the results are based on the entire population of each dictionary graph, no statistical tests were done. All differences would be highly significant because the number of words in each dictionary is so big. The effects themselves, however, are not very big; there are clearly many other factors underlying these variables apart from the dictionary latent structures.

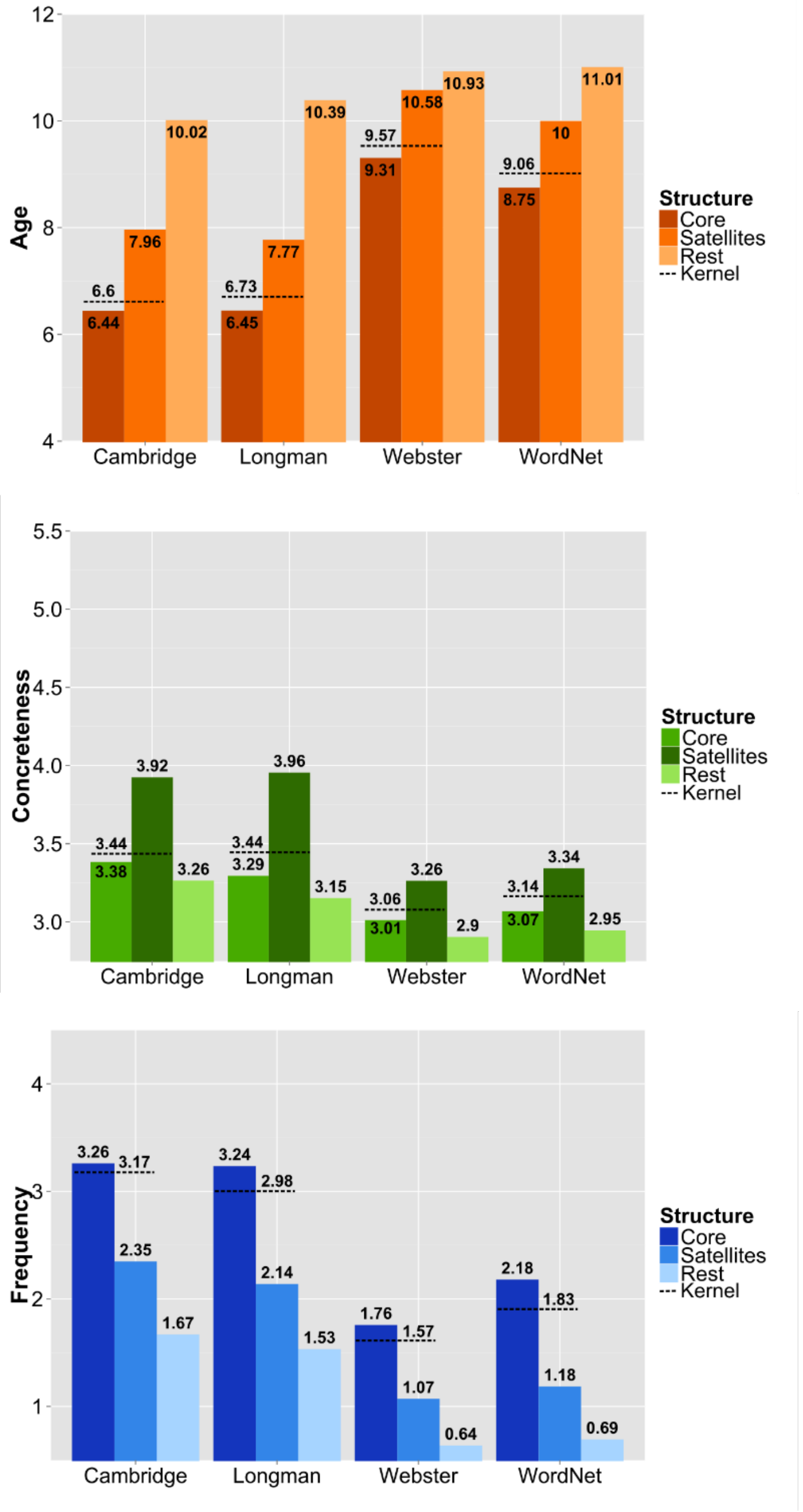

**Figure 3:** Average age, concreteness and frequency of words in Core, Satellites, Kernel and Rest. The pattern is the same for all four dictionaries: The Core is youngest and most frequent, then the Satellites, then the Rest. The Satellites are more concrete than the Core and the Rest, which are about equal (but see the gradients in Figure 7.)

The effect size for each of the pairwise differences in **Figure 3** is shown in **Figure 4**. Note that the biggest effect size tends to be for frequency. This may be because the psycholinguistic database coverage for frequency is close to 100% complete[5] for all the words in all three latent structures, Core, Satellites and Rest, whereas the coverage for age and concreteness declines with frequency, especially for the two larger dictionaries **(Figure 5)**. It is possible that the effect sizes for age and concreteness would have been larger, especially for the larger dictionaries, if the database coverage had been more complete. It is likely that the incompleteness of the data for age and concreteness is itself an indirect effect of word frequency: Age and concreteness data are lacking for the less frequent words[6].

All three variables – age, concreteness, and frequency – are intercorrelated (frequency/age: -0.5915; frequency/concreteness: 0.1583; age/concreteness: -0.3773). Decorrelating frequency from age and concreteness by recalculating effect sizes for only the residual variance left after removing the frequency variance reduces the effect sizes for age and concreteness (**Figure 6**). Age and concreteness data, which are much harder to gather than frequency data, are less available for less frequent words:

> "*From a list of English words that one of the authors (M.B.) is currently compiling, we selected all of the base words (lemmas) that are used most frequently as nouns, verbs, or adjectives*" (Kuperman et al 2012).
>
> "*Because ratings are only useful for well known words, we used a cut-off score of 85% known. In practice, this meant that not more than 4 participants out of the average of 25 raters indicated they did not know the word well enough to rate it. This left us with a list of 37,058 words and 2,896 two-word expressions (i.e., a total of 39,954 stimuli)*" (Brysbaert, Warriner & Kuperman 2014).

This introduces a frequency bias into our analysis, because of missing age and concreteness data for less frequent words. This frequency bias could either be (1) helping to reveal valid effects, (2) spuriously inflating them or (3) spuriously reducing them (**Figure 4**); or (4) removing the frequency bias by decorrelating frequency could be masking valid effects (**Figure 6**). We think it is unlikely that word frequency *causes* concreteness or age effects. It is more likely that age of acquisition and concreteness are part of the cause of frequency effects. But the direction of causality cannot be resolved by the available data.

---

[5] Words in our four dictionaries that had no values for SUBTELXus' frequencies were assigned frequency value zero. The SUBTELXus frequency data were collected on a corpus used as the reference database; zero means the word never occurred in that corpus.

[6] Brysbaert (personal communication) has noted that: "*Dictionaries contain many words not known to most humans. This is particularly the case for Webster and WordNet, of which nearly half the words refer to chemical or biological science words (types of plants, animals, tissues)… For many of these words there are no values for concreteness or Age of Acquisition. However, certainly for concreteness the main reason is that the raters do not know the words. Only experts are able to rate these.*"

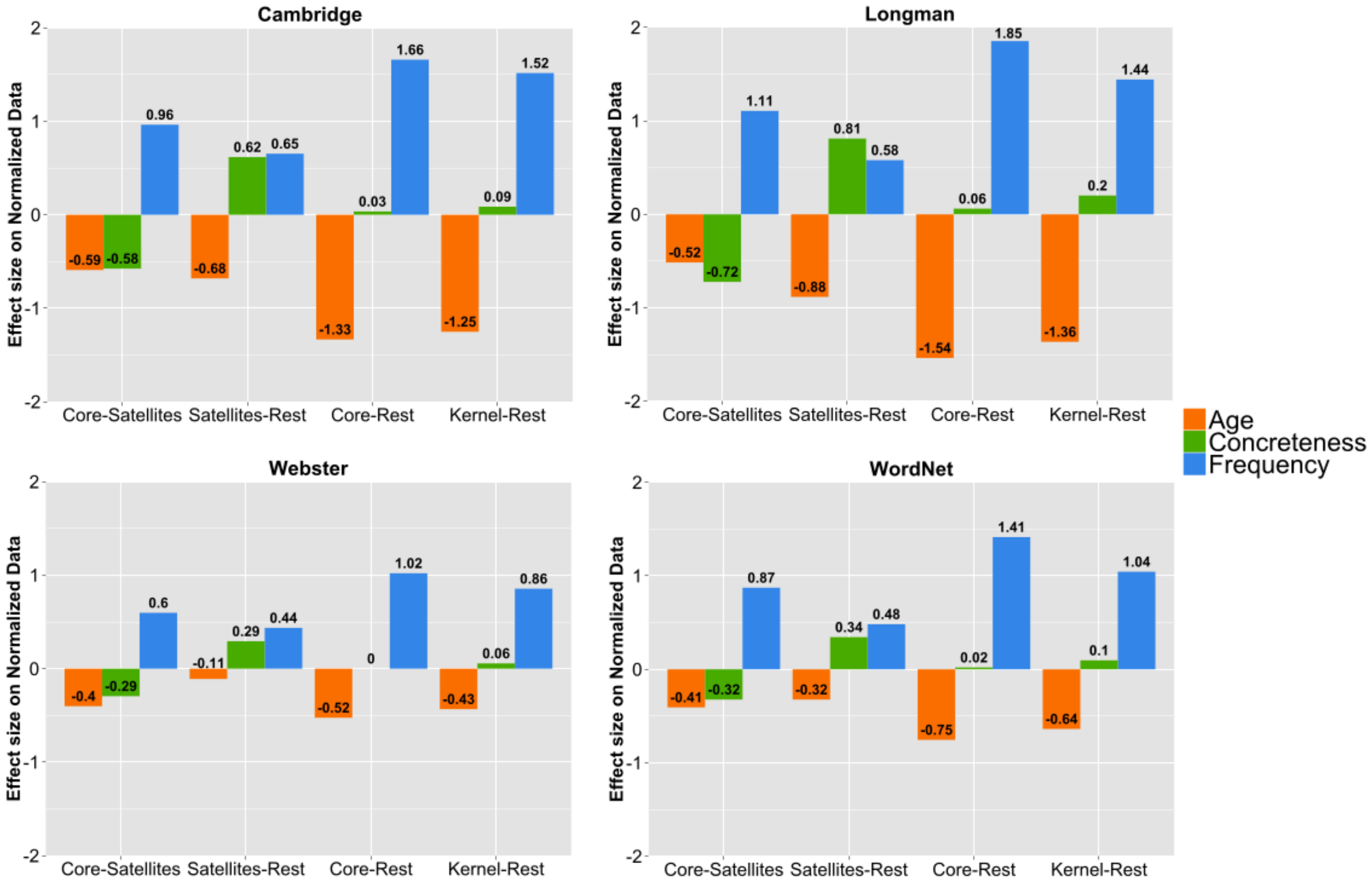

**Figure 4.** Effect size and direction for the principal comparisons among Core, Satellites, Kernel and Rest for age, concreteness and frequency, for each of the four dictionaries. Note that the effect size for frequency tends to be the largest, then age, then concreteness.

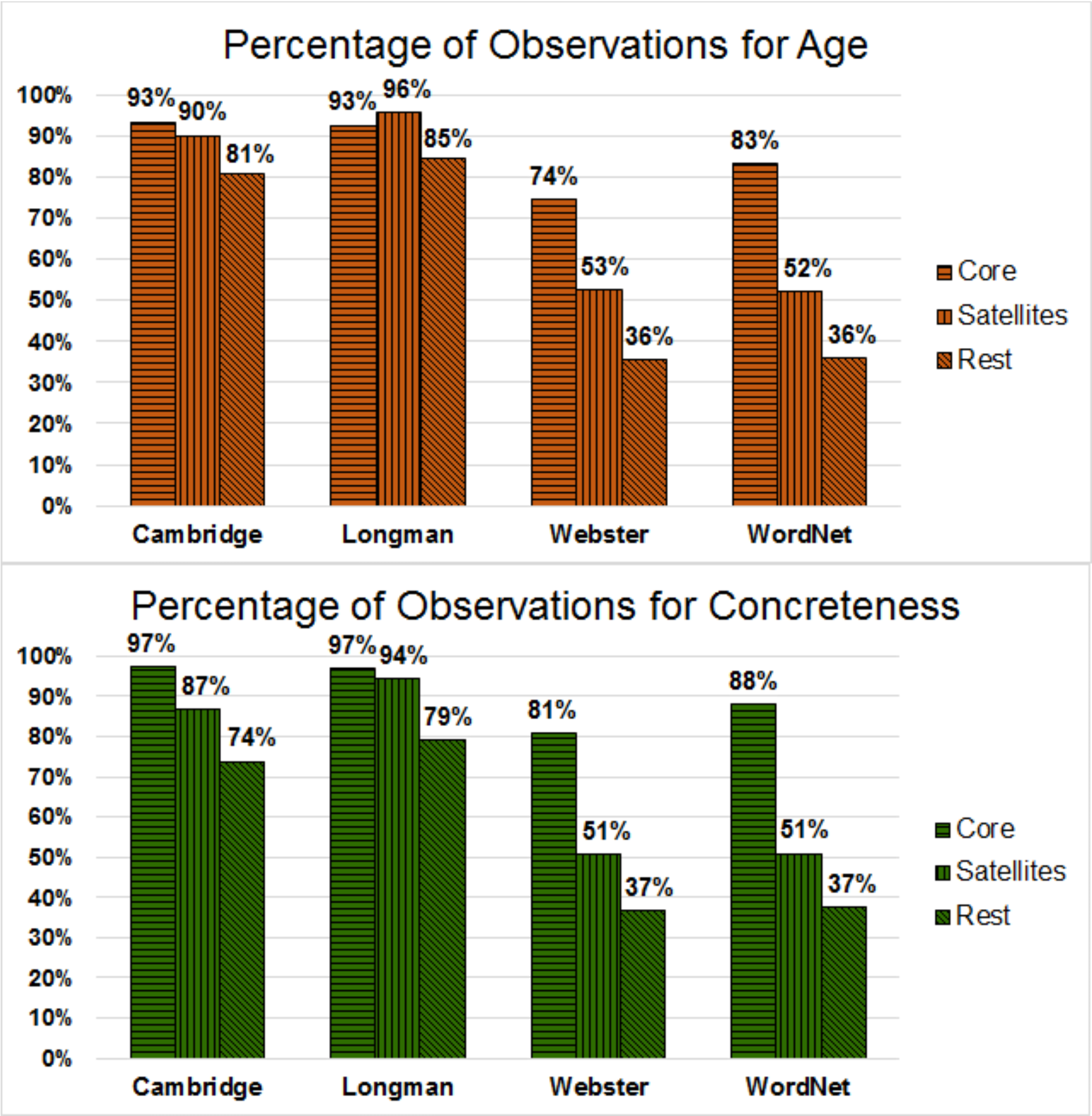

**Figure 5**. Percentage of words in Core, Satellites and Rest for which psycholinguistic data were available for age and concreteness for each of the four dictionaries. (Frequency data not shown because they are at 100% for all dictionaries.) Note that the percentage of available data is lower for the two bigger dictionaries, and decreases from the Core to the Satellites to the Rest.

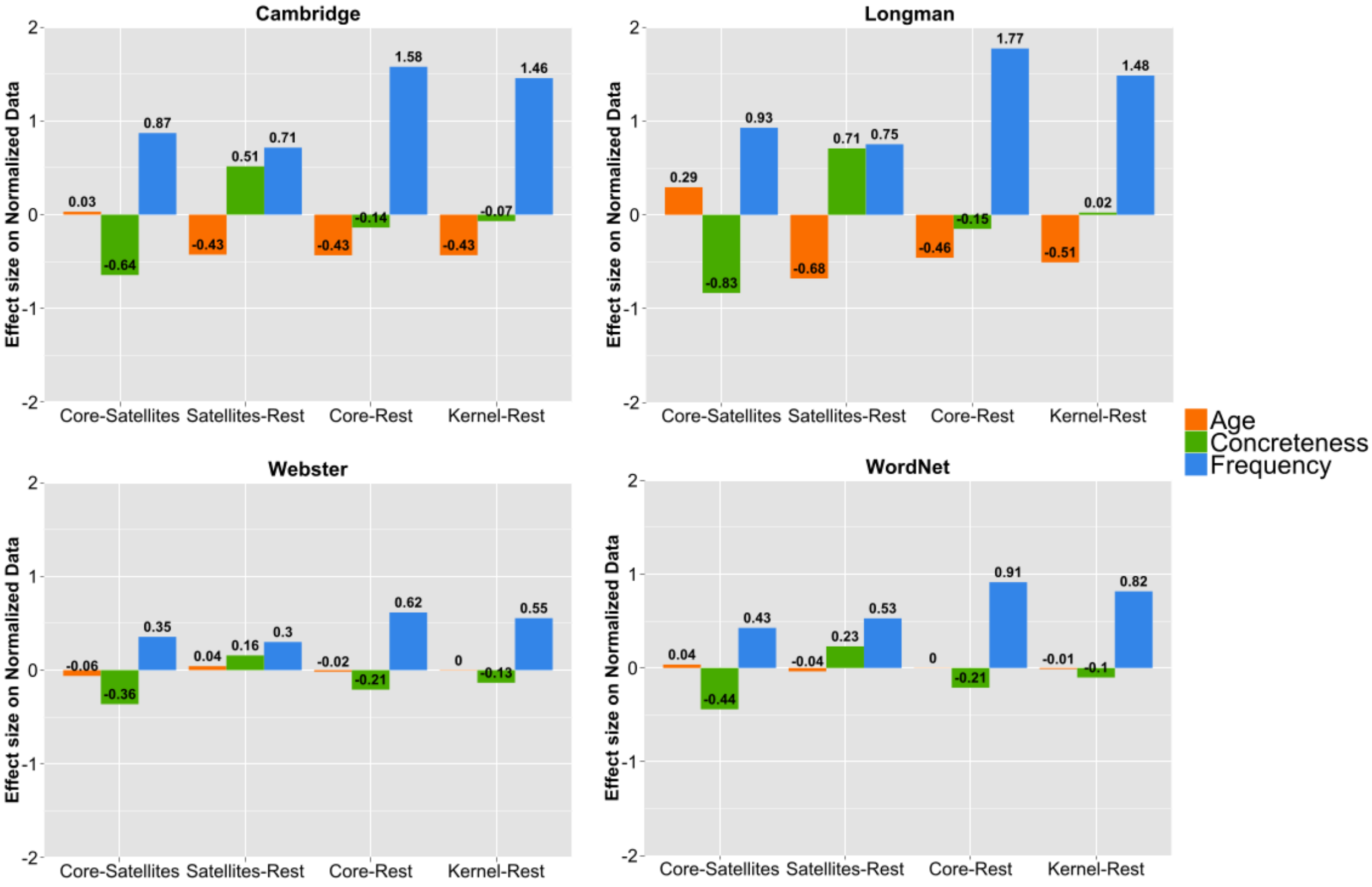

**Figure 6**. Effect size and direction for the principal comparisons among Core, Satellites, Kernel and Rest for age, concreteness and frequency, for each of the four dictionaries when correlation with frequency is removed. Age and concreteness effects are reduced considerably (cf. **Figure 4**), especially for the bigger dictionaries, in which the psycholinguistic database coverage for age and frequency was lower for the less frequent words (cf. **Figure 7**).

**Definitional Distance Gradients**. Alongside the main effects – the average differences in frequency, age and concreteness between the Core, Satellites and the Rest – our analysis also revealed two kinds of graded effects:

The upper part of **Figure 7** shows the gradient for the K-Hierarchy, which is the definitional distance from the Kernel to the words in the Rest of the dictionary (i.e. the number of definitional steps to reach a word starting from the Kernel). The first step in this gradient, from distance level 0 (the Kernel) to level 1 corresponds roughly to the main effects in **Figure 3**: For frequency there is a decrease from level 0 to 1 for all four dictionaries; then frequency is flat for all but Cambridge. For age there is an increase from level 0 to 1 (i.e., level 1 words are "older" – i.e., learned later – than the Kernel) for all four dictionaries, then descending slightly for all but WordNet. For concreteness there is a decrease (i.e., becoming more abstract) from 0 to 1, and then a gradual increase. Apart from the first step, from 0 to 1, the K-Hierarchy curves are hard to interpret because not only do the words at each succeeding distance level become fewer (**Table 1**) and less frequent, but the psycholinguistic database coverage for age (orange) and concreteness (green) is incomplete, especially for the two bigger dictionaries (**Figure 8**, left), many of whose rarer words are scientific, biological and technical terms.

The lower part of **Figure 7** shows the gradient for the C-Hierarchy, which is the definitional distance from the Core for words in the Satellite layer (i.e. the number of definitional steps to reach a Satellite word starting from the Core). Here the gradients are consistent for all four dictionaries and all three psycholinguistic variables: they are descending (less frequent) for frequency, rising (getting older) for age, and rising (getting more concrete) for concreteness. Here too the number of words diminishes at each distance level (**Table 2**), but for the two larger dictionaries there is a particularly marked decrease in database coverage for age (orange) and frequency (**Figure 8**, left). (This very visible negative correlation between definitional distance from the Core within the Satellite layer and psycholinguistic database coverage is probably due to the decline of word frequency with definitional distance from the Core within the Satellite layer (**Figure 7**, lower, blue). The red lines show the same effects when we analyze words that are present at the same level (intersection) in both large dictionaries (thick red line) and (separately) words that are present in both smaller dictionaries (thin red line).

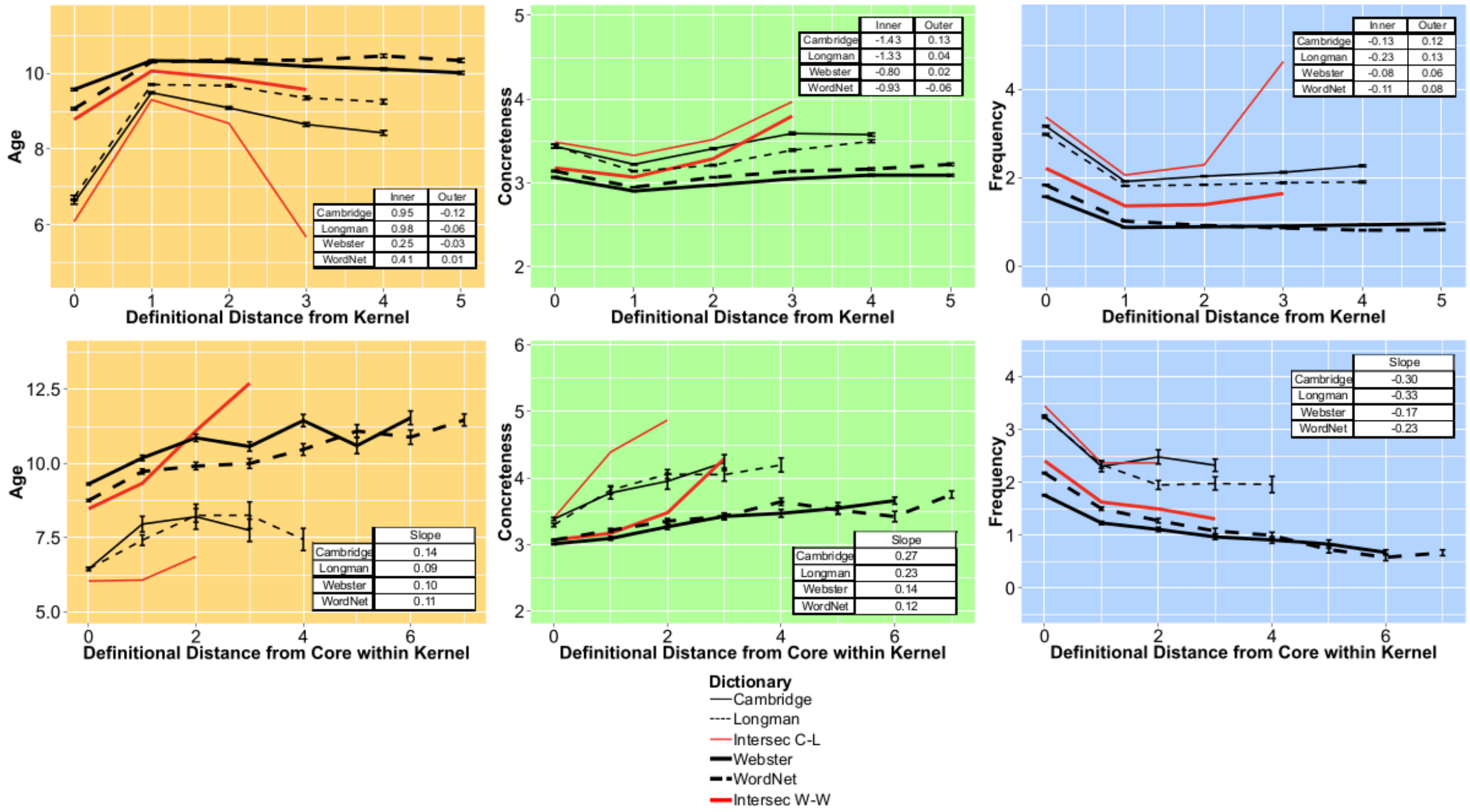

**Figure 7:** Average age, concreteness and frequency at each level of the definitional distance hierarchy starting from the Kernel through the Rest of the dictionary (K-Hierarchy, above), and within the Kernel, starting from the Core through the Satellites (C-Hierarchy, below), for each of the four dictionaries. K-Hierarchy: for age there is a big increase from the Kernel to level 1 and then a slight decrease at higher levels; for concreteness a slight decrease from K to 1, then slight increase; for frequency a big decrease from K to 1, then mostly flat. C-Hierarchy: increases for age and concreteness and decreases for frequency. All effects are stronger in the smaller dictionaries. The thick red lines show that the pattern is the same when considering only those words that occur at the same level (intersection) in both bigger dictionaries. The thin red lines show the pattern for words that occur at the same level in both smaller dictionaries.

| | K-Hierarchy | | | | | | | | | | | | | | |
|---|---|---|---|---|---|---|---|---|---|---|---|---|---|---|---|
| | 0 | 1 | 2 | 3 | 4 | 5 | 6 | 7 | 8 | 9 | 10 | 11 | 12 | 13 | 14 |
| Cambridge | 2241 | 15935 | 9483 | 4122 | 1663 (2796) | 730 | 276 | 105 | 20 | 2 | - | - | - | - | - |
| Longman | 2326 | 20555 | 12231 | 4966 | 1906 (3025) | 611 | 259 | 121 | 81 | 39 | 6 | 2 | - | - | - |
| Webster | 10955 | 59160 | 38111 | 19860 | 9577 | 4396 (7615) | 1904 | 666 | 292 | 201 | 89 | 38 | 25 | 2 | 2 |
| WordNet | 9802 | 48186 | 26186 | 12642 | 5379 | 2248 (4304) | 989 | 609 | 293 | 125 | 32 | 7 | 1 | - | - |

| | C-Hierarchy | | | | | | | | | | | | |
|---|---|---|---|---|---|---|---|---|---|---|---|---|---|
| | 0 | 1 | 2 | 3 | 4 | 5 | 6 | 7 | 8 | 9 | 10 | 11 | 12 |
| Cambridge | 2008 | 127 | 51 | 26 (54) | 20 | 4 | 4 | - | - | - | - | - | - |
| Longman | 1786* | 248 | 159 | 66 | 34 (64) | 14 | 6 | 4 | 4 | 2 | - | - | - |
| Webster | 7976* | 1220 | 640 | 425 | 245 | 153 | 106 (287) | 68 | 49 | 39 | 19 | 6 | - |
| WordNet | 6391 | 1270 | 683 | 443 | 308 | 252 | 179 | 117 (275) | 77 | 56 | 17 | 6 | 2 |

**Table 2**. Number of words at each level of the definitional distance hierarchy starting from the Kernel through the Rest of dictionary (K-Hierarchy, above), and, within the Kernel, starting from the Core through the Satellites (C-Hierarchy below), for each of the four dictionaries. Note that Figure 7 was truncated at the blue level past which frequencies became too low to be representative. Words past the truncation point were added to the blue value (total number of words for blue level shown in parentheses).

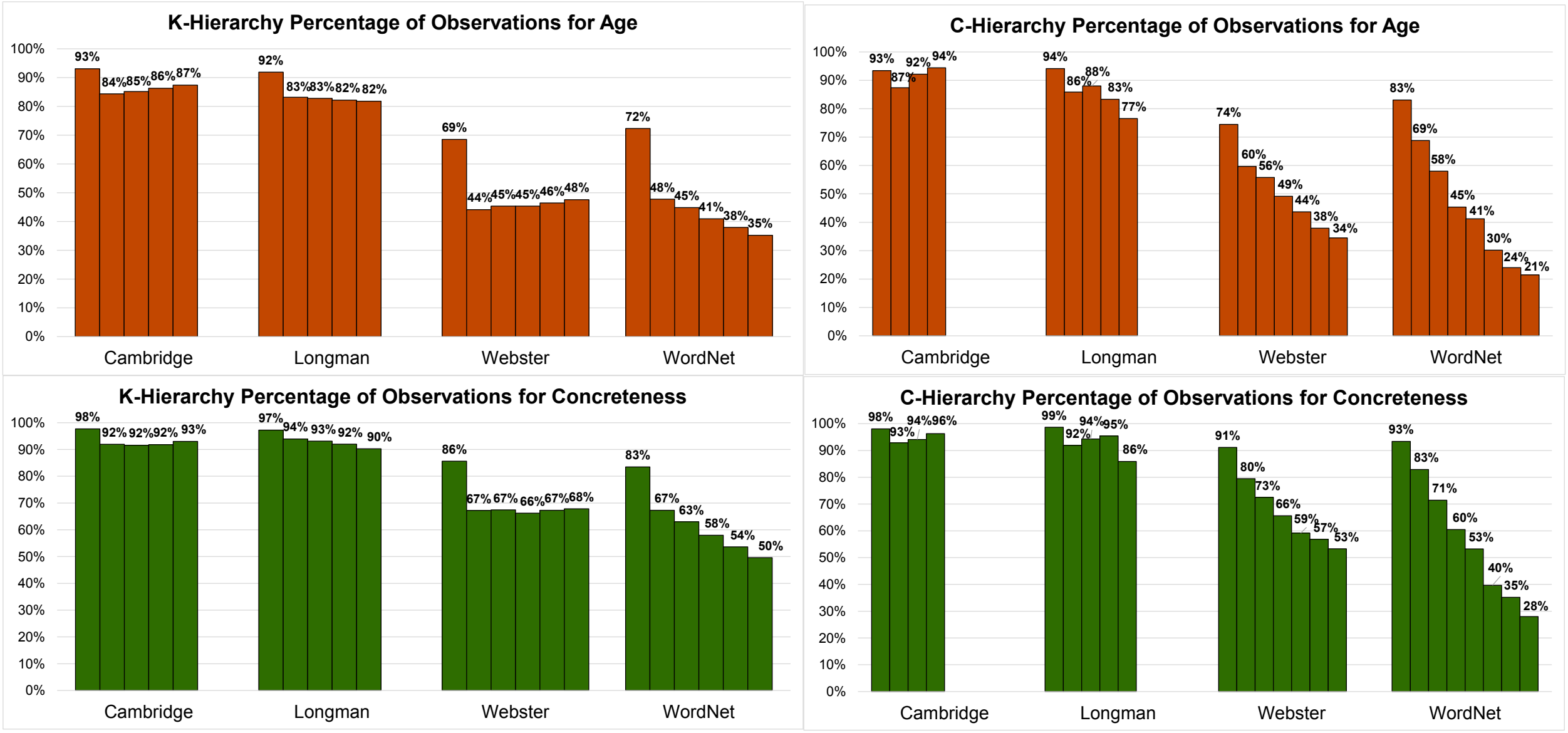

**Figure 8**. Percentage of words at each level of the definitional distance hierarchy starting from the Kernel through the Rest of dictionary (K-Hierarchy, left), and, only within the Kernel, starting from the Core through the Satellites (C-Hierarchy right), for which psycholinguistic data were available for age and concreteness for each of the four dictionaries. (Frequency data not shown because 100% for all dictionaries.) Note that the percentage of available data is lower for the two bigger dictionaries, and that within the Satellite layer it decreases with increasing definitional distance from the Core in the C-Hierarchy.

**Core and Satellite Components of the MinSets**. Because it takes so long to compute MinSets, even for the two small dictionaries, we do not have many of them yet; and for the two large dictionaries we so far only have one approximate MinSet each. Every MinSet is part-Core and part-Satellites. A natural question to ask is: What is the difference between the words in these two subcomponents of every MinSet? In the Kernel, the Core is more frequent, younger and less concrete than the Satellites. Comparing the words in the Core component of each MinSet with equal-sized random sets of Core words, and comparing the words in the Satellite component of each MinSet with equal-sized random sets of Satellite words also shows this ratio: For all four dictionaries, the Core component of the MinSet is more frequent, younger and less concrete than its random counterparts, and the Satellite component is less frequent, older and more concrete (**Figure 9**). (This effect was confirmed by t-tests ($p<0.001$) for the two smaller dictionaries, for which we had enough MinSets ($n=20$ and $n=19$ for Cambridge and Longman respectively). Because we were only able to compute one MinSet each for the two larger dictionaries, we could not do t-tests, but their pattern of results was the same as for the small dictionaries.) The Core/Satellite pattern is hence even more pronounced within the MinSets than in the Kernel as a whole.

Comparing the Core, Satellites and Rest in terms of parts of speech again points to the Satellite layer, which has more nouns and fewer adjectives, adverbs and verbs than the Core or the Rest in all four dictionaries **(Figure 10)**. This may be a hint of some sort of functional complementarity between Core and Satellites. Our digraphs and computations treat definitions as if they were just unordered strings of stemmatized content words' first meanings, ignoring syntax and even part of speech – yet definitions themselves are all subject/predicate propositions. The next step in further work will be to look into this formal black box, at the words themselves. There are very many words in the Core and the Satellites, and very many potential MinSets within each Kernel. To get a clearer idea of what the functional role of Core and Satellite words might be in making up a MinSet, in ongoing work we are examining the actual words themselves (rather than just their psycholinguistic correlates), as well as the actual definitions of which they are each a part, beginning with the words in individual mini-dictionaries generated by players in an online dictionary game (which will be briefly described in the next section of this paper).

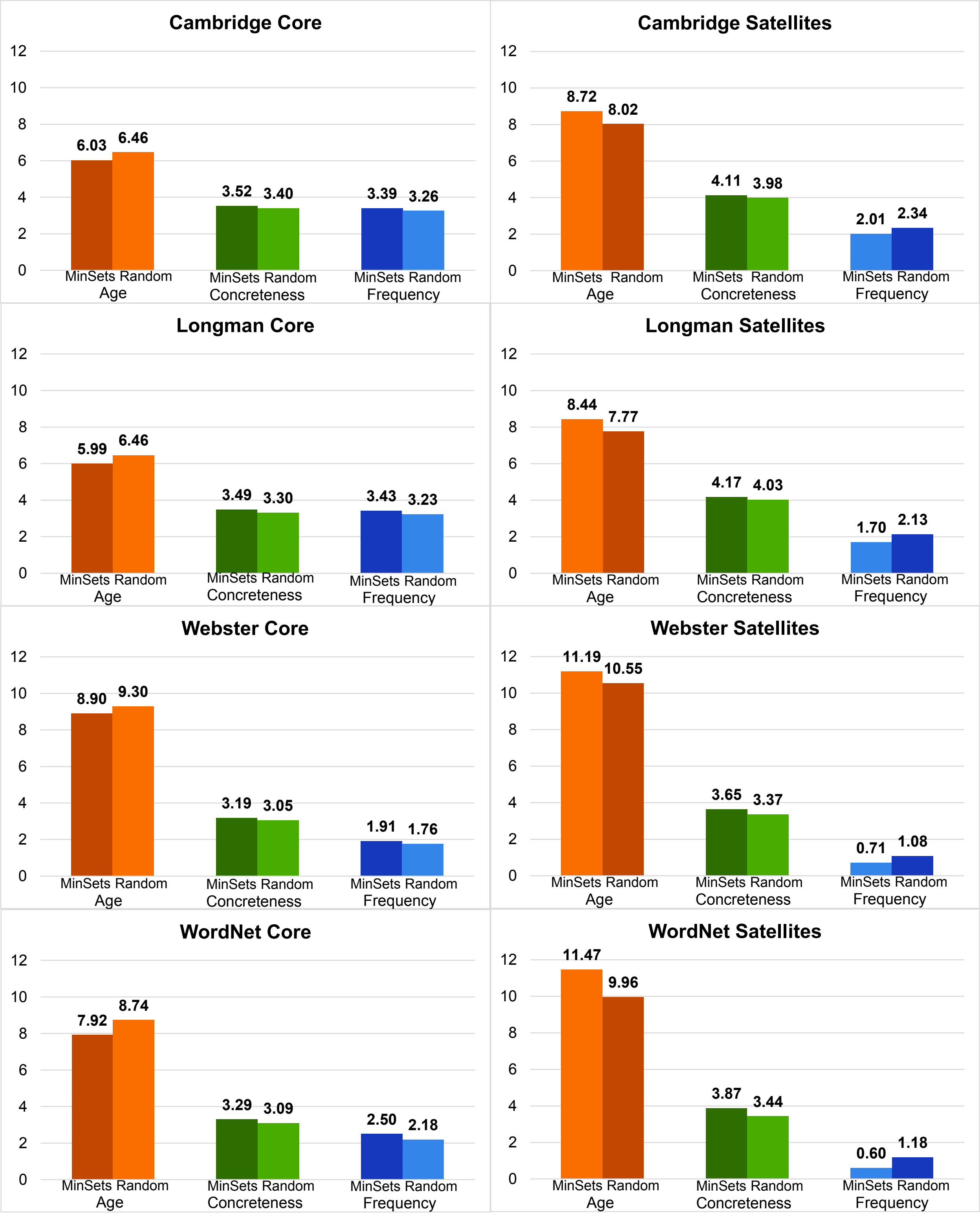

**Figure 9.** Comparing average age, concreteness and frequency of words in MinSets and equal-sized random subsets of the Core (left) and the Satellites (right) for each of the four dictionaries. In all four dictionaries the average MinSets are younger, more concrete and more frequent than random Core words and older, more abstract and less frequent than random Satellite words.

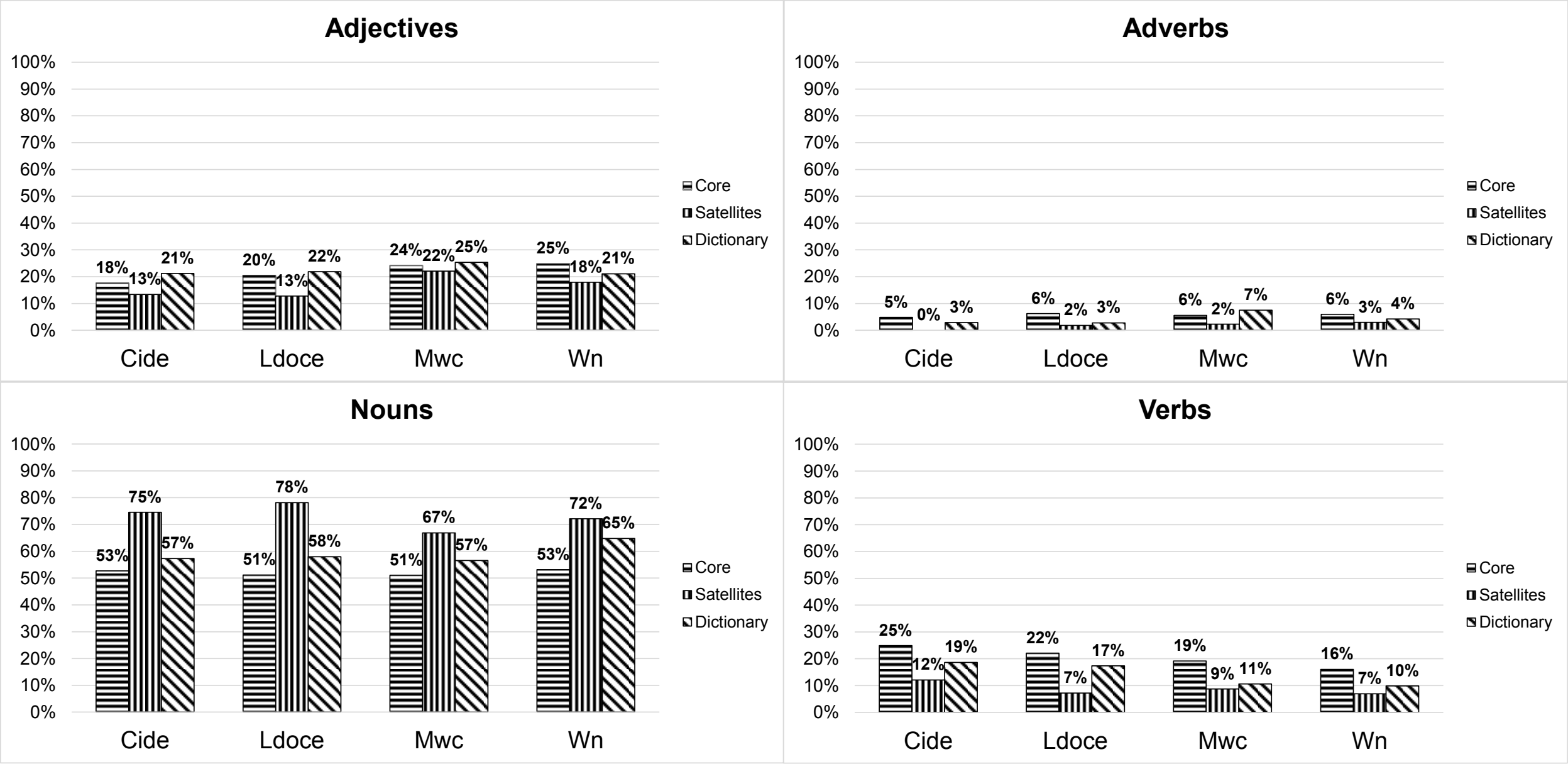

**Figure 10**. Percentage of parts of speech in the Core, Satellites and Rest for each of the four dictionaries. Note that the percentage of both nouns and adjectives is higher in the Satellite layer, whereas the percentage of verbs and adverbs is lower.

## Discussion

**Word-Association Graphs.** Graph-theoretic analysis has been used in other approaches to investigating how words are represented in the mental lexicon. Vitevitch (2008) has analyzed the phonological representations of words. Phonological representations are not semantic but share with dictionary definitions the property that they too are combinatory. In addition, the phonetic, phonemic and articulatory properties of words have been shown to influence word recognition and retrieval in the *lexical decision task* (Barber, Otten, Kousta & Vigliocco 2013). Variants of this task, in which subjects must quickly decide whether or not a series of letters is a word, have been widely used to study the representation of meaning in the mental lexicon (Aitchison 2012).

Word-association graphs have also been derived from a number of different measures of word-word association, including synonymy (Pilehvar & Navigli 2015), semantic similarity (De Deyne & Storms 2008; De Deyne, Navarro & Storms 2015), word co-occurrence frequency (e.g., Latent Semantic Analysis; Landauer & Dumais 1997; Jones, Willits, Dennis & Jones 2015) and people's free associations to words (Nelson, McEvoy & Schreiber 2004).

The relation between the definitional links in dictionary-definition graphs and the associative links in word-association graphs is ripe for further investigation. The underlying idea of an associative approach to the representation of meaning in the mental lexicon is that word meanings consist of word nodes and their pattern of interconnections in a network (Van Rensbergen, Storms & De Deyne 2015). The connections in a dictionary graph are *definitional*. There are defined words and defining words. The defining words point to the defined words. That relation is like a subject/predicate statement (although our analysis so far ignores syntax and function words, treating the defining content words as an unordered vector): The subject is simple (the word to be defined, e.g., "apple") and the predicate is compound and complex (the words that define the subject, e.g. "round red fruit").

The other important property peculiar to word-definition graphs is their *minimal grounding sets* (MinSets). An associative model of word meaning is ungrounded, in that there is no connection between the words and their referents; there are only connections between words and words. Words are just arbitrary symbols, and meaning is assumed to reside somehow in the words' patterns of interconnections. It is important to note, however, that *dictionaries (and dictionary graphs) are ungrounded too*, in that all they contain is words and the definitional links between them. A way in which this vicious circle of symbols can be broken, however, is *if the words in a MinSet* (actually or potentially – since there are many alternative potential Minsets, just as there are many alternative potential bases for an N-dimensional vector space) *are connected to their referents directly in some other way,* so that their meaning is *not* just a pattern of connections within the dictionary graph.

(Continuous N-dimensional vector spaces have an infinite number of potential bases, but N is always their minimal size. For a discrete, finite dictionary graph, its number of potential grounding MinSets is not infinite, but it is still very large – with N [the cardinality of the graph's minimum feedback vertex sets] being their minimal size.)

The size, N, of the MinSet is hence the smallest number of words that would need to be connected to their referents in this other way in order to generate meaning in what would then be a *dual-code* representation of word meaning in the mental lexicon. Our paper analyzes only the verbal (symbolic) code. The other code in which this verbal code must be grounded is a sensorimotor one, consisting of feature-detectors (inborn as well as learned) that can detect in their sensorimotor input the members of the category referred to by each (directly grounded) word (Harnad 1990, 2011; Barsalou 1999, 2010).

**Dual-Code Models of Mental Representation.** Dual-code models began with the work of Alan Paivio (1971,1986, 2014). The original idea was simply that we think in both words and pictures. But the connection between a picture and what it is a picture *of* is almost as ungrounded as the connection between a word and what the word is *about* – almost, because a picture can resemble its referent whereas (except for onomatompoeia) a word cannot. But resemblance is not enough for the perceptual recognition of the referent, especially when the referent is a category (e.g., "fruit") rather than an individual (this particular apple, "Alvin," now) and the category's members vary on many dimensions: The features that distinguish the members from the non-members can be complex and non-obvious.

What is needed in order to ground either pictures or words in their referents is a mechanism that can recognize sensorimotor categories. Modeling the capacity for the learning, recognition, and representation of sensorimotor categories has become a rich and fertile field (De Vega, Glenberg & Graesser 2008; Ashby & Maddox 2011; Lupyan 2012; Meteyard, Cuadrado, Bahrami & Vigliocco 2012; Pezzulo, Barsalou, Cangelosi, Fischer, McRae & Spivey, M. J. 2012; Maier, Glage, Hohlfeld & Rahman 2014; Folstein, Palmeri, Van Gulick & Gauthier 2015). In a dual-coding sensorimotor/symbolic model it is the sensorimotor module that needs to connect words to their referents. What our findings suggest is that *it need not connect them all, nor even most of them*: In principle (though not necessarily in practice), only the words in one MinSet would need to be connected directly to their referents by the sensorimotor module: all other words could then be formally encoded by the symbolic module via words alone, through the recombinatory expressive power of language, in the form of definitions, composed of those directly grounded MinSet words.

Steyvers & Tenenbaum's important study in 2005 found that word-association graphs tend to be small-world graphs consisting of a minority of highly connected central nodes fanning out into many less highly connected peripheral nodes, and that the words in the central nodes tend to be the ones that are more frequent and acquired earlier. They showed that as an associative net grows, the earlier nodes turn into the ones with many connections. This pattern would fit the growth of an individual speaker's vocabulary across time or the growth of the vocabulary of a language across time.

Our analysis of definitional depth in a dictionary-graph – from the periphery to the Kernel to the Satellites to the Core, with each MinSet turning out to be part-Satellite and part-Core – shows that the words in a dictionary-graph become increasingly frequent and young with increasing depth as we move inward toward the dictionary's Core (**Figure 2**). This finding parallels Steyvers & Tenenbaum's finding that words are increasingly frequent and younger toward the central nodes in their associative graphs. Our interpretation that our grounding MinSets are learned earlier is consistent with Steyvers & Tenenbaum's interpretation that their central words are learned earlier. Later studies will examine the extent to which the same words are involved in both kinds of graph.

As noted, a definitional network can be more than just an associative network: If it is also one of the two interconnected modules in a dual-coding sensorimotor/symbolic representation of meaning in the mental lexicon then *the internal representation of word meaning itself is more than just associative*. The difference resides in the functional role of a MinSet in a dictionary graph. Each MinSet is a (potential) generator of the meaning of all the rest of the words in the dictionary, via formal definitional connections alone. Hence, in principle, none of the rest of the words need to be grounded directly in their referents: They can instead inherit their grounding indirectly, via the formal definitional connections – *on condition that the words in the generator MinSet have been grounded by the sensorimotor module.* The meanings of the words in the MinSet need to have been learned previously via direct sensorimotor experience rather than via definition. That is why those words are younger: not just because they were learned first but because they acquired their meaning in a fundamentally different way.

Of course it is not only unlikely but almost certainly false that it will turn out to be as neat and simple as the following: First (1) one MinSet of word meanings of size N is learned from direct inductive experience (internally implemented by sensorimotor feature-detectors that allow each category member to be reliably assigned to its category and hence to the word that names it). Then (2) every other content word in the language, and the category it refers to, is learned indirectly via purely verbal definitions consisting of re-combinations of the directly grounded words of the MinSet (along with any further words reached from them indirectly via definition).

What is more realistic than (1) and (2) is to assume that the growth of the mental lexicon continues to be hybrid sensorimotor/symbolic throughout a speaker's lifespan, with some later direct sensorimotor experience complementing and reinforcing the acquisition of new meanings alongside verbal instruction whenever it is needed or makes learning and understanding easier.

But what cannot be hybrid is the *initialization* of this dual representational system: Some words first have to be grounded directly through sensorimotor experience; then those words can go on to ground further word learning, either via whatever new words they can define symbolically by recombination, or via hybrid sensorimotor/symbolic learning.

Semantic and syntactic "bootstrapping" occurs when children learn the meaning of words from written or spoken context (Gleitman & Landau 1994). Dictionaries simply *define* words: all words. Being *told* in words what a word refers to (just as being shown, by pointing, what a word refers to) is importantly different from inferring, from context, what the word might refer to (bootstrapping), though the two may work hand in hand: Dictionaries sometimes provide examples of the use of the defined word in context, but that is by way of *supplementing* the definition: the user is not expected to "bootstrap" to the meaning from just the example of its use in context.

What really comes first, before words or even language itself, is *categories themselves*. A category need not be given a name. To categorize is simply to *do* the right thing with the right *kind* of thing (Cohen & Lefebvre 2005). With language, the right thing to do might be to name (or describe) the category; but pre-verbally it can be anything that needs to be *done* with the members of the category, and *not-done* with the non-members: approaching, avoiding, eating, mating-with, manipulating, etc., based on the sensorimotor features (affordances) that afford doing the right thing (Montesano, Lopes, Bernardino & Santos-Victor 2008; Yürüten, Şahin & Kalkan 2013).

So in our view the mental lexicon is itself hybrid – a dual-code representational system consisting of learned sensorimotor feature (affordance) detectors for the grounding words (and any later hybrid words) plus re-combinatory and purely symbolic (i.e., verbal) definitions and descriptions for the referents of the words that are learned via words alone. No doubt the words in such a dual-code mental lexicon will have associative properties too, based on similarities in sound, meaning and context of use, but meaning itself cannot consist of word-word associations all the way down, for the same reason it cannot consist of words all the way down. Linguistic symbols have to be grounded in their real-world referents.

**Frequency, Concreteness and Age of Acquisition.** Frequency is the most objective of the three psycholinguistic variables we have tested so far, but it is also the least specific (and its measurement is complicated by factors such as the type/token distinction, polysemy and familiarity). Age of acquisition (AoA) is much more specific, but less objective: It is very hard to get data on the age at which children actually first hear and understand particular words, whereas retrospective adult estimates about this could be influenced by many complicating factors. Yet AoA nevertheless seems to be a reasonably reliable parameter.

Frequency and AoA are highly correlated (younger words are more frequent), so one of the big challenges is to try to disentangle them and interpret the difference. Brysbaert, Van Wijnendaele & De Deyne (2000) showed that age effects are more sensitive to the semantic (meaning) properties of words than to their word-form frequency (lexical or phonological) properties, with younger-learned words being processed faster for meaning than older-learned words. This accords with the symbol-grounding interpretation of our dictionary findings: The Kernel is younger than the words in the rest of the dictionary; and within the Kernel, the Core is youngest, with words getting younger and younger the shorter their definitional distance from the Core. Brysbaert's finding that the speed of processing of a word's meaning is influenced by how early it was learned rather than just how frequent it is in the language reinforces our finding of the primacy of the Satellite and Core words in the encoding of meaning in dictionaries. It also accords with Steyvers & Tennenbaum's (2005; Tenenbaum, Kemp, Griffiths & Goodman 2011) finding that the earliest nodes of a semantic network are the most strongly connected ones.

The question of the causal role of frequency is still an open one. There is no doubt that frequency is correlated with grounding – the Core words are the most frequent ones, then the Satellites, then the Rest. The frequency gradient within the Satellite layer also follows this pattern, and there is no detectable frequency gradient in the rest of the dictionary, even though the frequency database is 100% complete. It may well be that some words are learned earlier because they are more frequent in the language: But why are they more frequent in the language? Frequency is undeniably the strongest of the psycholinguistic correlates of the latent structures of the dictionary. But its causal role must be explained by something other than frequency: It cannot be frequency all the way down, any more than it can be definitions all the way down.

Concreteness/abstractness is more problematic because it can mean so many different things. At bottom, it can mean the difference between physical objects, like tables, plants and animals, and their properties, such as hard, green or fast. Properties are more abstract than the objects of which they are the properties, but properties can themselves have properties, such as texture, color or speed, and so on, in a rising hierarchy of higher-order properties. If higher-order properties were its only dimension, abstractness/concreteness would be straightforward (except for parts, composite properties, and intransitive properties), but another dimension of concreteness is the sensorimotor one: the closer something is to something that can be perceived directly with the senses, the more concrete we consider it. That means that "shape," a 3rd order property (apple, round, shape), is more concrete than "quark," a 1st-order object (but imperceptible). Moreover, both objects and properties of any order can have emotional associations (reflected in other neural correlates as well), which would again make them more concrete (in the sensorimotor sense), no matter how abstract they are (in the property-hierarchy sense):

The lexical decision task (of deciding whether or not a string of letters is a word) had at first seemed to indicate that concrete words are recognized faster than abstract ones. But Kousta and coworkers have found that when other variables are controlled, certain abstract words are recognized faster because they have emotional associations (Kousta, Vigliocco, Vinson, Andrews & Del Campo 2011; Barber, Otten, Kousta & Vigliocco 2013; Vigliocco, Kousta, Della Rosa, Vinson, Tettamanti, Devlin & Cappa 2014; cf. Paivio 2013). This raises the question of what it is that we mean by "concrete" vs. "abstract" (Borghi & Binkofski (2014). As noted, *abstract* has at least two distinct senses: (1) an ontic, object-centered sense, in which individual physical objects are concrete and their hierarchy of higher-order properties becomes increasingly abstract; and (2) a sensorimotor, subject-centered sense, in which both objects and properties become more concrete the more palpable they or their properties are to our senses. In this second sense, emotions and emotional associations, being sensory, would be concrete rather than abstract.

Kousta and co-workers' finding that (some) abstract words can reverse the reaction time advantage of concrete over abstract words in the lexical decision task because of their emotional connotations may well be related to the multidimensionality of our judgments of concreteness/abstractness. What is not clear yet is the relationship between performance in the lexical decision task and how meaning differences between words (as opposed to the difference between words and non-words) are perceived or mentally represented. The C-hierarchy of definitional distance we found radiating from the Core outward through the Satellite layer may help cast some light on how meaning differences are represented, because it is in this hierarchy that the concreteness effect seems to be concentrated. But this will only become testable in the individual mini-dictionaries that are currently being generated in our dictionary game, rather than the full-sized lexicographers' dictionaries, which are a composite product originating from the many individual mental lexicons of many different individual lexicographers.

**Other Psycholinguistic Variables.** Any evidence that can distinguish word frequency from the other psycholinguistic variables can be important in trying to make causal interpretations. Comparing word frequency data with data in which people simply indicate whether or not they know a word will be very useful (Keuleers, Stevens, Mandera & Brysbaert 2015; Brysbaert, Stevens, Mandera & Keuleers, in press). Brysbaert calls this new variable, which correlates only .50 with word frequency, *word prevalence* ("knowledge of the word in the crowd") and suggests (personal communication) that "one would expect all words of the Core to be known, whereas the Satellites may contain more specialized knowledge."

If, as is likely, the familiarity of a word depends on the expertise or some other characteristic of sub-populations, that too needs to be taken into account (Vinson, Ponari & Vigliocco 2014; Della Rosa, Catricalà, Vigliocco & Cappa 2010). Dictionaries tend to be generic, but there is scope for investigating specialized technical dictionaries, glossaries and "ontologies" separately in their own right as well as for using psycholinguistic data derived from their corresponding expert sub-populations rather than generic psycholinguistic data. This pertains to human performance studies rather than to graph-theoretic analyses of dictionaries per se. It is also likely to become increasingly important in our work on the dictionary game.

Another promising new variable based on estimating *how* a word was learned (whether from direct sensorimotor experience or verbally) rather than just *when* it was learned (although the two are of course correlated) is Mode-of-Acquisition (Della Rosa, Catricalà, Vigliocco & Cappa, 2010; Connell & Lynott (2012), Dellantonio, Mulatti, Pastore & Job 2014; Brysbaert, Warriner & Kuperman 2014). Sensorimotor experience is sensory as well as motor; it includes anything that is experienced – i.e. felt – via any modality, whether the usual five "external" senses or the interoceptive ones, such as proprioception, kinesthesia or emotion (Barsalou, Simmons, Barbey & Wilson 2003; Montesano, Lopes, Bernardino & Santos-Victor 2008; Frak, Nazir, Goyette, Cohen & Jeannerod 2010; Guan, Meng, Yao & Glenberg 2013; Vigliocco, Kousta, Della Rosa, Vinson, Tettamanti, Devlin & Cappa 2014). Mode-of-acquisition may well prove to be at least as relevant to our dictionary analyses as the three variables we have analyzed, but it is unfortunately not yet available for enough words to be useable in the present study.

## Conclusions

What we have learned from the graph-theoretic analysis of dictionaries so far is that someone who knows the meaning of a grounding set of as few as 373 words – for a small dictionary of 25,132 words (first meanings only) or 1396 words for a larger dictionary of 91,388 words (first meanings) – could in principle learn the (first) meanings of all the rest of the words in the dictionary through definition alone. It does not follow, of course, that that is the way we actually do learn the meanings of all the rest of the words. If the grounding set was learned through direct sensorimotor experience, it is probable that a lot of later words are learned in a hybrid way, through a combination of direct experience and verbal definition (or description, or instruction or explanation). Most of our categories are not lexicalized at all, and are described (rather than defined) by ad hoc verbal descriptions: there is no dictionary entry for "things that are bigger than a breadbox," for example, nor for "things that I saw last Tuesday" – nor even, in most people's vocabularies, for "feeling glee at another's misfortune." But even the words in ad hoc verbal descriptions of unlexicalized categories have to be grounded, just as dictionary definitions have to be. So that's back to the grounding set.

**Language and Propositions.** One can equally well ask: "Why couldn't the meanings of *all* words be learned through direct sensorimotor grounding?" First of all, if it really were possible to learn the meaning of every word through direct sensorimotor experience, then why bother to have words at all? Presumably it is to transmit what one of us has learned (say, via direct experience) to another who has not. Here we cannot avoid considering the question of the nature of language itself, and its adaptive value for our species. No other species speaks (in any modality, including gesture). What do other species lack, and what has ours gained, for having evolved the capacity for language? It is the ability to say anything and everything that can be said: the ability to express every possible proposition.

Most of what we say (even questions and commands) consists of subject/predicate propositions. Some propositions are “deictic,” which means that they point to the immediate sensorimotor here-and-now: “She is here.” Deictic terms are all function words, which were excluded from our dictionary analysis. We were only interested in content words, which, as noted, are the names of categories, and make up almost all the words in the dictionary. “Apples are red” is a simple subject/predicate proposition, very much like most of what we say (including this very sentence). The proposition could be formalized as stating that “apples” are a member (or a subset) of “things that are red.” It could be the reply to someone asking either “What color are apples?” or “What things are red?” This already has most of the features of asking the question “What does ‘apple’ mean?” or “What is an ‘apple’?” and then learning from an interlocutor or from a dictionary that (to a first approximation) “An apple is a round, red fruit.”

So far, all these categories (“apple, “red,” “fruit”) could have been learned either from a verbal description/definition or from direct sensorimotor experience: They are all pretty concrete, and they could all be learned early, fairly quickly, and without any particular risk. “Goodness,” “truth, and “beauty” are becoming more abstract – although “that’s good (true, beautiful)” and “that’s not good (true, beautiful)” could be learned from positive and negative examples through direct experience too. Learning what “quark”or “quiddity” mean nonverbally, from direct experience, would be quite a bit harder, and the meaning of “peekaboo-unicorn” (“a one-horned horse that vanishes without a trace whenever either senses or an instrument are aimed at it”) would be impossible to learn directly via the senses, whereas its verbal definition is just as well grounded as the definition of apple.

**Category Learning: the Hard and Easy Way.** Now suppose the category that someone lacks is not “apples” but “toadstools,” and that the person is starving, and the only thing available to eat is edible mushrooms or poisonous toadstools that look very much like the edible mushrooms. Being told, by someone who knows, that “The striped gray mushrooms are poisonous toadstools” could save someone a lot of time (and possibly their life) by making it unnecessary to find out through direct trial-and-error experience which kind is which.

And that, in a nutshell, is our hypothesis about the nature and adaptive value of language (Blondin Massé et al 2013): Language makes it possible to learn new categories by word of mouth, by recombining already grounded category names into propositions defining/describing new categories, instead of having to learn them the hard way, from direct experience. But to make it possible to learn by word of mouth, some words, at least, still have to be grounded in direct experience. The grounding would need to occur earlier, before the grounded words could be used to define and transmit further categories. And because grounding is sensorimotor, the grounding words would tend to be more concrete.[7]

There is no reason to expect the grounding words to be unique and identical for everyone. The minimal grounding set of any individual's mental lexicon might be like the basis set of an N-dimensional vector space: the basis can generate every point in the vector space, but it is not unique: just a set of N linearly independent points with the property that linear combinations of them can generate any and every point in the vector space. But, because people share a lot of common experiences, and because this is in turn reflected in the vocabulary of their language, there is nevertheless reason to expect that some words will be part of many people's grounding vocabularies; so those words would be spoken and written more frequently (see the frequency curves as well as the red curves for intersections in **Figure 7**).

This hypothesis is certainly not *entailed* by our findings on the greater frequency of Kernel words, the earlier age of acquisition of Core words, the greater concreteness of Satellite words, or the multiplicity of MinSets. But if the hypothesis were correct, it would help make sense of some of these findings: Not all. It remains a puzzle why Kernel words in the Satellite layer become increasingly concrete but also older and less frequent, the greater their definitional distance from the Core. We will not understand that until we get a better idea of the complementary role of Core words and Satellite words in making up a MinSet. But even for our two smallest dictionaries there are still very many words in their Kernels (over 2000), and they have very many different MinSets (each of c. 400 words each). So we are currently also generating tiny dictionaries by means of an online dictionary game:[8]

The participant is given a word, asked to define it, and then to define the words used to define it, and so on, until all the words used have been defined. This yields dictionaries with an average size of about 200 words, 90% of them in the Kernel, and with MinSets of about 30 words, 2/3 of them Satellite words and 1/3 Core words (which is a reversal of our observed ratio for the full-size dictionaries) (**Table 1**). We hope that these much smaller dictionaries generated by individuals will turn out to reflect the way meanings are represented in the mental lexicon and will allow us to get a better idea of the complementary roles played by Core and Satellite words in jointly making up a MinSet. We also hope that this article will encourage "crowd-sourcing" the analysis of dictionary graphs for further psycholinguistic variables as well as in further languages[9].

---

[7] As noted, however, there are some conceptual problems with the notion of concreteness (Borghu & Binkofski 2014), and hence also with judgments of concreteness: To name a kind, like "apple," rather than just a unique individual on a unique occasion, is already to abstract.

[8] Readers are invited to try their hand at generating a few mini-dictionaries at http://lexis.uqam.ca:8080/dictGame. It will deepen their understanding and appreciation of the nature of a dictionary and their own individual mental lexicon as well as many of the points discussed in this paper.

[9] We have archived the Kernels (separated into Satellites and Core) for all four dictionaries online for use by other researchers at (URL to come).

**Language evolution.** Symbol grounding concerns not only the origin of the words acquired during a speaker's lifetime but also the origin of words themselves, in the evolution (both biological and historical) of language (Pagel, Atkinson & Meade 2007; Monaghan 2014). There may turn out to be a relation between word origins and the psycholinguistic correlates of Core, Satellite and MinSet words. Language may have begun concretely, with sensorimotor categorization, pantomime and gesture (Steklis & Harnad 1976; Harnad 2011; Cangelosi & Parisi 2012; Blondin Massé, Harnad, Picard & St-Louis 2013) and some of this may still be reflected in language acquisition and vocabulary today. But that question is still speculative and awaits the outcome of dictionary analyses across other languages.

What is certain is that it is human language that has not only generated the biggest and richest digital database on the planet, far bigger than what our species could have generated by sensorimotor means alone; but it is also language (of which computation is just a subset) that has generated the means to derive meaning from all those data – on condition that those minimal means, at least, are grounded in the kinds of sensorimotor experience that we share with all other species on the planet.